\documentclass{article} % For LaTeX2e
\usepackage{iclr2027_conference,times}

\usepackage{amsmath,amsfonts,bm}

\def\eqref#1{equation~\ref{#1}}
\def\1{\bm{1}}

\DeclareMathAlphabet{\mathsfit}{\encodingdefault}{\sfdefault}{m}{sl}
\SetMathAlphabet{\mathsfit}{bold}{\encodingdefault}{\sfdefault}{bx}{n}

\usepackage{hyperref}
\usepackage{url}

\usepackage{multirow}
\usepackage{algorithm}
\usepackage{algpseudocode}
\usepackage{booktabs}
\usepackage{dsfont}
\usepackage{graphicx}
\usepackage{caption}
\usepackage{amsmath,amssymb}
\usepackage{xcolor}
\usepackage{threeparttable}
\usepackage{makecell}
\usepackage[table]{xcolor}

\definecolor{myred}{HTML}{C92321}
\definecolor{mygreen}{HTML}{56BA77}
\colorlet{myredbg}{myred!8}
\colorlet{mygreenbg}{mygreen!10}
\definecolor{groupgray}{HTML}{F2F2F2}

\definecolor{rkl}{RGB}{30,56,107}
\definecolor{fkl}{RGB}{240,145,72}

\title{Divergence-Routed On-Policy\\Distillation for Language Models}

\iclrfinalcopy
\author{
Tiantian Peng$^{1}$ \quad
Guannan Lv$^{3}$ \quad
Hongyang Li$^{1}$ \quad
Qiuhe Hong$^{1}$ \quad
Hang Cheng$^{2}$ \\
Hongjian Dou$^{3}$ \quad
Xiangyu Wu$^{3}$ \quad
Yuyang Liu$^{1}$\thanks{Corresponding author.} \\
$^{1}$Peking University \\
$^{2}$Tsinghua University \\
$^{3}$Kuaishou Technology
}

\begin{document}

\maketitle

\pagestyle{fancy}
\fancyhead{}  % 清空原页眉
\fancyhead[L]{Under review as a conference paper at ICLR 2027}
\thispagestyle{fancy}

% 后

\begin{abstract}
On-policy distillation (OPD) transfers knowledge from a teacher model to a student along student-generated trajectories, typically by minimizing the reverse Kullback--Leibler (RKL). However, RKL can provide weak corrective signals for teacher-supported tokens to which the student assigns very low probability. This raises two questions: which positions are most consequential for the behavioral gap between the student and teacher, and what additional supervision is needed at these positions? Through cross-sampling, we find that divergence-based routing more effectively pinpoints positions contributing to the teacher--student performance gap than entropy-based. Further analysis reveals teacher-mode under-coverage at high Jensen--Shannon (JS) divergence positions, where standard OPD increasing the large teacher--student probability gaps. Motivated by this findings, we propose \textbf{Divergence-Routed On-Policy Distillation (DrOPD)}. DrOPD retains RKL supervision at all positions, and selectively adds forward Kullback--Leibler (FKL) at positions with the high JS divergence. Experiments across six math benchmarks show that DrOPD improves over standard OPD and entropy-based routing. DrOPD also reduces the severe under-coverage cases by 37.9\% relative to OPD, supporting targeted FKL as an effective complement to OPD. 

\end{abstract}

\section{Introduction}
\label{sec:introduction}

On-policy distillation (OPD) provides an effective way to transfer knowledge from a stronger teacher to a student by providing supervision along student-generated trajectories
~\citep{agarwal2023onpolicy,gu2024minillm,lu2025onpolicydistillation}, thereby reducing the mismatch between offline distillation supervision and the student's inference-time state distribution.
Standard OPD is commonly implemented by minimizing the reverse Kullback--Leibler (RKL)  ~\citep{kullback1951information} between the student and teacher next-token distributions at every response position.
However, RKL is weighted by the student probability, and can provide weak corrective signals for teacher-supported tokens that already receive very low student probability.
As a result, some teacher modes may remain under-covered despite improvements in overall alignment.

Recent work has explored selective token-level supervision, often using the student or teacher entropy to identify important positions
~\citep{jin2026entropy,zhang2026tailoring,hou2026uni,xu2026tip,ko2026scaling,liu2026teacher}.
Entropy, however, measures uncertainty within a single model, but uncertainty does not necessarily indicate a consequential difference between student and teacher. For example, both models may be uncertain while assigning similar probabilities to the available candidates. 
This motivates two questions: \emph{which positions are most consequential for the teacher--student performance gap, and what additional supervision is needed at these positions?}

We use controlled \emph{cross-sampling}~\citep{meng2026sparse} to identify routing signals that effectively locate positions contributing to the teacher--student performance gap. The student serves as the default generation policy, while the teacher supplies tokens only at positions selected by a routing criterion. We compare Jensen--Shannon (JS) divergence~\citep{lin1991divergence}, RKL, and forward Kullback--Leibler (FKL) against student entropy, teacher entropy, and random routing. The results reveal that divergence-based strategies achieve similar intervention efficiency and outperform others. On AMC23, divergence-based strategies reach teacher performance with teacher intervention at fewer than 10\% of generated positions. These results identify teacher--student disagreement as an effective signal for locating positions where additional guidance matters most.

We then examine the probability distributions at high-JS positions and find pronounced teacher-mode under-coverage: on the same contexts, the student assigns substantially lower probabilities than the teacher to the recorded tokens. Standard OPD training further exacerbates severe under-coverage at these positions. This behavior is consistent with the gradient structure of RKL: for a candidate with fixed positive teacher probability, the RKL logit gradient vanishes as the student probability approaches zero, whereas FKL retains a teacher-directed corrective signal. These observations motivate adding FKL supervision at high-disagreement positions to help recover teacher-supported tokens that RKL struggles to correct.

Based on this diagnosis, we propose \textbf{Divergence-Routed On-Policy Distillation (DrOPD)}. DrOPD retains RKL supervision at all response positions and adds FKL supervision to the fraction $\rho$ of positions with the highest teacher--student JS divergence. This design preserves the standard OPD learning signal while directing complementary mode-covering supervision toward the positions identified by our diagnosis. Experiments across six mathematical reasoning benchmarks show that DrOPD improves mean Avg@8 over standard OPD by 1.84\%  for the 1.7B--4B student--teacher pairs. It also outperforms entropy-based routing methods, supporting disagreement as a more effective guide for allocating additional supervision. On fixed high-JS contexts, standard OPD increases the fraction of positions with severe probability gaps from 3.83\% in the base student to 4.48\%. DrOPD reduces this fraction to 2.79\%, a relative reduction of 37.9\% compared with OPD. These results connect the performance gains to improved recovery from teacher-mode under-coverage.

Our contributions are threefold:
\begin{itemize}
\item \textbf{Identifying where additional supervision matters.} Controlled cross-sampling shows that teacher--student divergence, RKL, FKL and JS divergence, locates behaviorally important positions more effectively than single-model entropy.
\item \textbf{Diagnosing severe teacher-mode under-coverage.} Fixed-context analysis reveals severe probability gaps at high-JS positions that become more frequent after standard OPD, motivating complementary FKL supervision through gradient analysis.
\item \textbf{Divergence-Routed On-Policy Distillation.} We propose DrOPD, which combines all-position RKL with selectively allocated FKL,  improving reasoning performance across multiple settings and and reducing severe under-coverage.
\end{itemize}

\section{Diagnosing Token Importance in On-Policy Distillation}
\label{sec:token_importance}

We first use controlled cross-sampling to evaluate signals for identifying positions where teacher intervention improves downstream performance, then examine teacher-mode under-coverage at high-disagreement positions and its connection to KL gradients. Given a prompt $x$ and prefix $y_{<t}$, let $p_t$ and $q_t$ denote the student and teacher next-token distributions, respectively, \textbf{renormalized over the union of their top-16 candidate sets}. We compare single-model uncertainty, measured by student entropy $H(p_t)$ and teacher entropy $H(q_t)$, with teacher--student disagreement, measured by RKL $D_{\mathrm{KL}}(p_t\|q_t)$, FKL $D_{\mathrm{KL}}(q_t\|p_t)$, and JS divergence: $ D_{\mathrm{JS}}(p_t,q_t)=\tfrac12 D_{\mathrm{KL}}(p_t\|m_t)+\tfrac12 D_{\mathrm{KL}}(q_t\|m_t) $, where $m_t=\tfrac12(p_t+q_t)$.

\begin{figure}[t]
    \centering
    \includegraphics[width=1.0\linewidth]{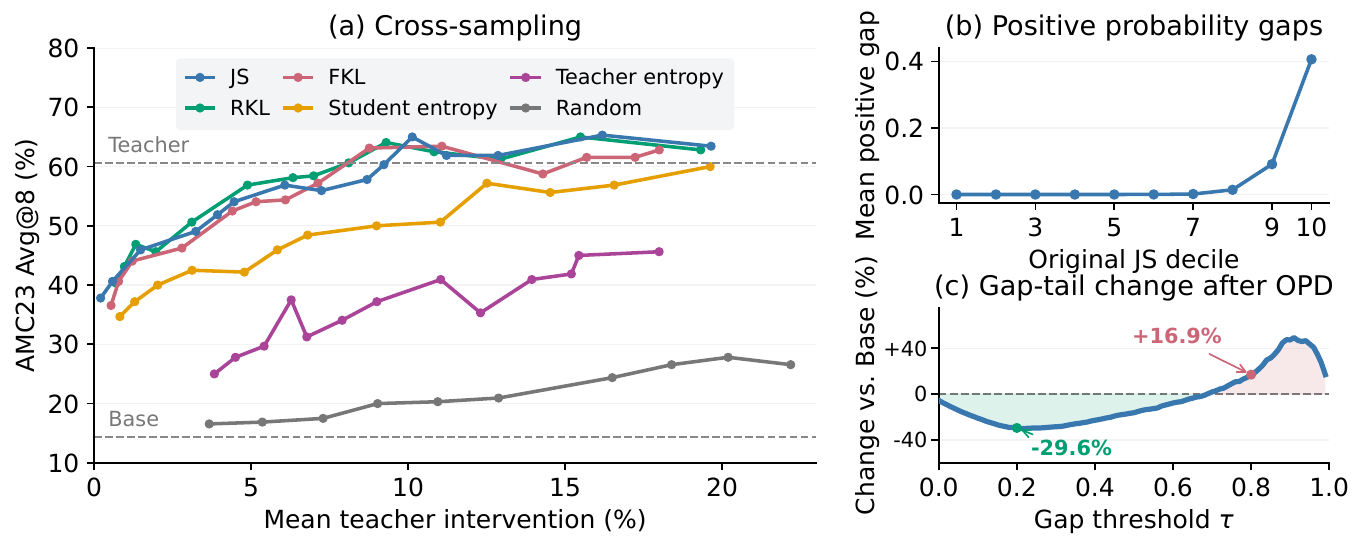}
    \caption{\textbf{Cross-sampling and teacher-mode under-coverage at high-JS positions.} (a) AMC23 accuracy versus teacher-intervention rate; dashed lines indicate standalone Teacher and Base performance. (b) Mean positive probability gap across original JS deciles, computed over recorded tokens with positive gaps. (c) Relative change from Base to OPD in the fraction of positions with gaps at or above each threshold, evaluated on fixed original JS $>0.06$ positions.}
\label{fig:cross_sampling_combine}
\end{figure}

\subsection{Controlled Cross-Sampling}
\label{sec:cross_sampling}
Following \citet{meng2026sparse}, we use controlled cross-sampling to assess how effectively routing signals identify positions that benefit from teacher intervention. At each decoding step, giving a routing score $r_t$ and threshold $\tau$, we define $S_t=\mathbb{I}[r_t>\tau]$ and sample
\begin{equation} 
y_t\sim\begin{cases} q_t, & S_t=1,\\ p_t, & S_t=0. \end{cases} 
\label{eq:cross_sampling} 
\end{equation} 
The sampled token is appended to the prefix. We compare routing based on JS divergence, RKL, FKL, student entropy, and teacher entropy, together with a random baseline. For each stategy, we independently sweep the threshold $\tau$; for random routing, we manually change the intervention probability. Each setting is evaluated on 40 AMC23\citep{li2024numinamath}  problems, with eight responses per problem. We measure the actual teacher-intervention ratio as:
\begin{equation}
\eta=\frac{1}{N}\sum_{i=1}^{N}\frac{1}{T_i}
\sum_{t=1}^{T_i}S_t^{(i)}, \qquad N=320,
\label{eq:intervention_ratio}
\end{equation}
where $T_i$ is the length of response $i$. We compare Avg@8 as a function of $\eta$: higher accuracy at comparable intervention ratios indicates that a routing signal more effectively identifies positions where teacher guidance improves task performance.

\subsection{Divergence Identifies Behaviorally Critical Positions}
\label{sec:divergence_critical}

Figure~\ref{fig:cross_sampling_combine}a shows that divergence-based routing uses teacher intervention more efficiently than others. JS divergence, RKL, and FKL achieve similar performance, reaching near-teacher accuracy with intervention at fewer than 10\%. For example, JS routing achieves 60.31\% Avg@8 with 9.23\% intervention, close to the teacher's 60.63\%. In contrast, student- and teacher-entropy routing achieve only 50.00\% and 37.19\%, respectively. These results suggest that divergence is an effective signal for locating behaviorally critical positions, while entropy alone is less informative because a model can be uncertain even when the student and teacher make similar predictions. 

We use JS divergence for the remaining analysis and the routing mechanism in Section~\ref{sec:method} because it is symmetric and bounded. Having identified high-disagreement positions is valuable, we next examine their probability distributions to understand what additional supervision they need.

\subsection{From High Disagreement to Teacher-Mode Under-Coverage}
\label{sec:undercoverage}
\paragraph{Measuring under-coverage on fixed contexts.}
We examine the probability distributions at high-JS positions using 320 Base--Teacher cross-sampled responses. To quantify teacher-mode under-coverage, we define the positive teacher--student probability gap as 
\begin{equation} 
\delta_t=\bigl[q_t(v)-p_t(v)\bigr]_+, \qquad [a]_+=\max(a,0),
\label{eq:prob_gap} \end{equation} 
where $v$ is the recorded token at position $t$. A larger $\delta_t$ indicates that the student assigns less probability to candidate $v$ than the teacher does.

Figure~\ref{fig:cross_sampling_combine}b shows that under-coverage is more pronounced at high-JS positions. For the base student, the mean positive gap increases from 0.014 in the eighth JS decile to 0.406 in the highest. Thus, positions identified as valuable intervention sites also exhibit substantial deficits in the probability assigned to teacher-supported tokens. Standard OPD reduces many probability gaps but exacerbates severe under-coverage. Figure~\ref{fig:cross_sampling_combine}c shows the relative change from Base to OPD in the fraction of fixed high-JS positions with $\delta_t\geq\tau$, where $\tau$ is the gap threshold. At $\tau=0.2$, this fraction decreases by 29.6\%. At $\tau=0.8$, however, it increases from 3.83\% to 4.48\%, a relative increase of 16.9\%. Both comparisons use the same positions with original JS $>0.06$.

\paragraph{Understanding under-coverage through gradient analysis.}
The persistence of severe under-coverage is consistent with the gradient structure of RKL. For a student distribution $p$ parameterized by logits $z(v)$ and a fixed teacher distribution $q$, the logit gradients are 
\begin{align} 
\frac{\partial D_{\mathrm{KL}}(p\|q)}{\partial z(v)} &= p(v)\left(\log\frac{p(v)}{q(v)}-D_{\mathrm{KL}}(p\|q)\right),
\\ 
\frac{\partial D_{\mathrm{KL}}(q\|p)}{\partial z(v)} &= p(v)-q(v). 
\label{eq:kl_gradients} 
\end{align} 
For fixed $q(v)>0$, the RKL gradient vanishes as $p(v)\to0$, weakening the direct corrective signal for severely under-covered candidates. The FKL gradient instead approaches $-q(v)$, so gradient descent continues to increase the student's logit for these candidates. This complementary gradient behavior motivates adding FKL supervision at high-disagreement positions.

\begin{figure}[t]
    \centering
    \includegraphics[width=1.0\linewidth]{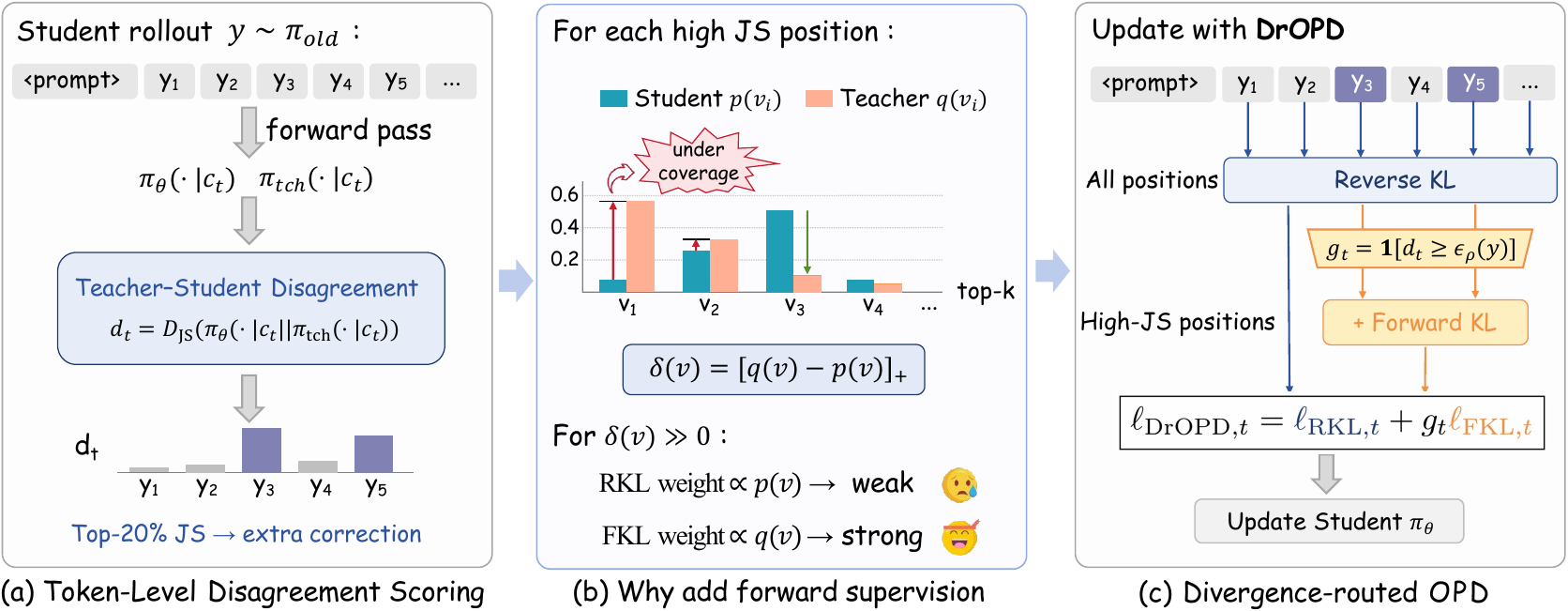}
    \caption{\textbf{Overview of DrOPD.} (a) We compute teacher--student JS divergence at each rollout position and select the top-20\% with the highest disagreement. (b) High-JS positions can exhibit under-coverage of teacher-supported modes, for which RKL provides weak corrective signals. (c) DrOPD retains RKL supervision at all positions and adds FKL supervision at the high-JS positions.}
    \label{fig:dropd}
\end{figure}

\section{Divergence-Routed On-Policy Distillation}
\label{sec:method}

Motivated by the above findings, we propose DrOPD, which retains the RKL throughout student-generated trajectories and adds FKL at selected high-JS positions. Figure~\ref{fig:dropd} illustrates the method.

\subsection{Disagreement Scoring and Position Selection}
\label{sec:drop_routing}

Set a student policy $\pi_{\mathrm{old}}$ to generate student responses. At position $t$, let $c_t$ denote the generated prefix. We form the union candidate set from the student's and teacher's top-$k$ candidates, and renormalize the two distributions over the union candidate: :
\begin{equation}
\mathcal C_t=\operatorname{TopK}_k(p_t)\cup\operatorname{TopK}_k(q_t),\quad
\widetilde p_t(v)=\frac{p_t(v)}{\sum_{u\in\mathcal C_t}p_t(u)},\quad
\widetilde q_t(v)=\frac{q_t(v)}{\sum_{u\in\mathcal C_t}q_t(u)}.
\label{eq:candidate_distributions}
\end{equation}
This union retains both student-preferred candidates and teacher-supported candidates. We use $k=16$ and score disagreement at each position by $d_t=D_{\mathrm{JS}}(\widetilde p_t,\widetilde q_t)$.
Let $y$ denote the collection of student responses in the current batch, and let $\epsilon_\rho(y)$ be the $(1-\rho)$-quantile of disagreement scores over its valid response positions. We select high-disagreement positions using
\begin{equation} 
g_t=\mathds{1}\!\left[d_t\geq\epsilon_\rho(y)\right]. 
\label{eq:drop_router} 
\end{equation}
We set $\rho=0.2$, allocating additional supervision to the top 20\% highest-disagreement positions.

\begin{table*}[t]
    \centering
    \setlength{\tabcolsep}{3pt}
    % \small
    \caption{\textbf{Mathematical reasoning results (Avg@8, \%).} Both panels select and average over all six benchmarks. Results marked with $\dagger$ are reproduced by us.}
    \label{tab:main_result}
    \begin{tabular}{lcccccc>{\centering\arraybackslash}p{1.7cm}}
\toprule

Method
& AIME24
& AIME25
& AMC23
& Minerva
& Olympiad
& MATH500
& Mean(6) \\
\midrule
\rowcolor{groupgray}
\multicolumn{8}{c}{\small\textit{\textbf{Baselines}}} \\
1.7B-Base &  0.42 & 1.25 & 14.38 & 6.34 & 8.18 & 23.88 & 9.07 \\
4B-Base & 10.00 & 5.42 & 30.00 & 6.48 & 24.78 & 40.38 & 19.51 \\
4B-GRPO &  24.17 & 17.50 & 60.63 & 30.19 & 48.53 & 83.10 & 44.02 \\
8B-GRPO  & 25.83 & 21.67 & 68.13 & 35.29 & 51.52 & 86.00 & 48.07 \\

\addlinespace[1pt]
\midrule
\rowcolor{groupgray}\multicolumn{8}{c}{\small\textit{\textbf{Student: 1.7B-Base}  \ \ \ \        \textbf{Teacher: 4B-GRPO}}} \\

OPD & 12.92 & 8.33 & 45.31 & 21.23 & 33.03 & 68.63 & 31.57 \\
Stu-entropy  & 12.08 & 7.08 & 47.19 & 24.59 & 34.55 & 70.43 & 32.65 \\
Tch-entropy  & 12.92 & 7.92 & 46.25 & 23.44 & 34.77 & 71.00 & 32.72 \\
EOPD$^{\dagger}$~\citeyearpar{jin2026entropy} & 11.25 & 7.50 & 45.94 & 23.07 & 34.72 & 69.93 & 32.07 \\
\rowcolor{mygreen!12} DrOPD & 14.17 & 11.67 & 46.56 & 24.03 & 33.88 & 70.15 & 33.41 \\

\addlinespace[1pt]

% ============== Stu: 4B / Tch: 8B ==================
\midrule
\rowcolor{groupgray}\multicolumn{8}{c}{\small\textit{\textbf{Student: 4B-Base}\ \ \ \    \textbf{Teacher: 8B-GRPO}}} \\

OPD  & 24.58 & 18.33 & 61.88 & 33.00 & 47.92 & 82.15 & 44.64 \\
Stu-entropy & 23.33 & 19.17 & 59.38 & 32.86 & 47.61 & 82.93 & 44.21 \\
Tch-entropy & 20.83 & 19.17 & 62.19 & 33.64 & 47.57 & 82.53 & 44.32 \\
EOPD$^{\dagger}$~\citeyearpar{jin2026entropy}  & 22.08 & 18.75 & 62.19 & 32.67 & 48.02 & 83.15 & 44.48 \\
\rowcolor{mygreen!12} DrOPD  & 25.00 & 19.17 & 63.44 & 32.54 & 48.07 & 83.18 & 45.23 \\
\bottomrule
\end{tabular}

\end{table*}

\subsection{All-Position RKL with Routed FKL Supervision}
\label{sec:routed_fkl}
Using $\widetilde p_t$ and $\widetilde q_t$, we define the top-$k$ approximate distillation loss at position $t$ as 
\begin{equation}
\ell_t(\theta)
=
\sum_{v\in\mathcal C_t}
\widetilde p_t(v)\log\frac{\widetilde p_t(v)}{\widetilde q_t(v)}
+
g_t\sum_{v\in\mathcal C_t}
\widetilde q_t(v)\log\frac{\widetilde q_t(v)}{\widetilde p_t(v)}.
\label{eq:drop_composition}
\end{equation}
The RKL preserves the standard OPD objective at every response position. When $g_t=1$, the additional FKL supplies teacher-weighted supervision to help correct under-covered candidates.

In practice, we optimize a policy-gradient surrogate using the following distillation reward:
\begin{equation}
R_t(v)
=
-\widetilde p_t(v)\log\frac{\widetilde p_t(v)}{\widetilde q_t(v)}
+
g_t\widetilde q_t(v),
\qquad v\in\mathcal C_t.
\label{eq:drop_advantage}
\end{equation}
The first term supplies the standard OPD signal, while the second adds the FKL correction. We use these rewards directly as detached advantages,
$A_t(v)=\operatorname{sg}[R_t(v)]$,
and update the student by maximizing the PPO-style clipped surrogate:
\begin{equation}
\mathcal J(\theta)
=
\frac{1}{|\mathcal B|}
\sum_{t\in\mathcal B}
\sum_{v\in\mathcal C_t}
\min\!\left[
w_t(v;\theta)A_t(v),
\operatorname{clip}\!\left(
w_t(v;\theta),1-\epsilon_{\mathrm{clip}},1+\epsilon_{\mathrm{clip}}
\right)A_t(v)
\right].
\label{eq:drop_ppo}
\end{equation}
Here, $w_t(v;\theta)$ is the importance-sampling weight, $\mathcal B$ contains the valid response positions in the batch, and $\epsilon_{\mathrm{clip}}$ is the clipping parameter. The teacher remains frozen during each policy update.

% \begin{table*}[t]
%     \centering
%     \setlength{\tabcolsep}{2.5pt}
%     \small
%     \input{tab/ablation}
% \end{table*}

\section{Experiments}
\label{sec:experiments}

\subsection{Experimental Settings}
\paragraph{Models and training.}
Our primary pair uses Qwen3-1.7B-Base as the student and Qwen3-4B-Base-GRPO as the teacher. We also evaluate a Qwen3-4B-Base student with a Qwen3-8B-Base-GRPO teacher~\citep{yang2025qwen3,shao2024deepseekmath}. We use verl\citep{sheng2024hybridflow} as our training framework. Students train on DAPO-Math-17K\citep{yu2026dapo} for one epoch with four student-generated responses per prompt, with batch size of 64, and a 7,168-token response limit. DrOPD uses the union of both models' top-16 candidates and routes added FKL to the top-20\% JS positions. All diagnostic and sensitivity analyses use the primary model pair.

\paragraph{Evaluation.}
We report AIME24, AIME25~\citep{balunovic2026matharena}, AMC23~\citep{li2024numinamath}, Minerva~\citep{lew2022minerva}, Olympiad~\citep{he2024olympiadbench}, and MATH500~\citep{hendrycks2021measuring}. Each problem has eight responses sampled at temperature 0.7 and top-$p=0.95$, with a 7,168-token limit. Avg@8 measures mean response correctness.

\paragraph{Compared methods.}
We compare DrOPD with standard OPD and entropy-based supervision strategies. \textbf{OPD} applies RKL supervision at all response positions. \textbf{Stu-entropy} and \textbf{Tch-entropy} retain all-position RKL and add FKL at the top 20\% of positions ranked by the corresponding entropy. We also reproduce \textbf{EOPD}~\citep{jin2026entropy}, which adds FKL at positions where teacher entropy exceeds $0.8$ while retaining RKL supervision at all positions.

\subsection{Main Results}
\label{sec:main_results}
Table~\ref{tab:main_result} shows that DrOPD improves over standard OPD on both model pairs. For the 1.7B--4B pair, mean Avg@8 increases from 31.57\% to 33.41\%, a gain of 1.84 percentage points, with improvements on all six benchmarks. For the 4B--8B pair, DrOPD raises the mean from 44.64\% to 45.23\% and improves five of the six benchmarks. These results show that the benefit of targeted FKL supervision extends across both student scales. 

DrOPD also outperforms entropy-based routing for both model pairs. For the 1.7B--4B pair, stud-entropy routing, tch-entropy routing, and the EOPD achieve 32.65\%, 32.72\%, and 32.07\%, respectively, compared with 33.41\% for DrOPD. This comparison highlights the value of the routing criterion. Consistent with the cross-sampling results, teacher--student disagreement provides a more effective guide for allocating forward supervision than either model's uncertainty alone.

\subsection{Training Dynamics}
\label{sec:training_dynamics}

\begin{figure}[t]
\centering
\includegraphics[width=\linewidth]{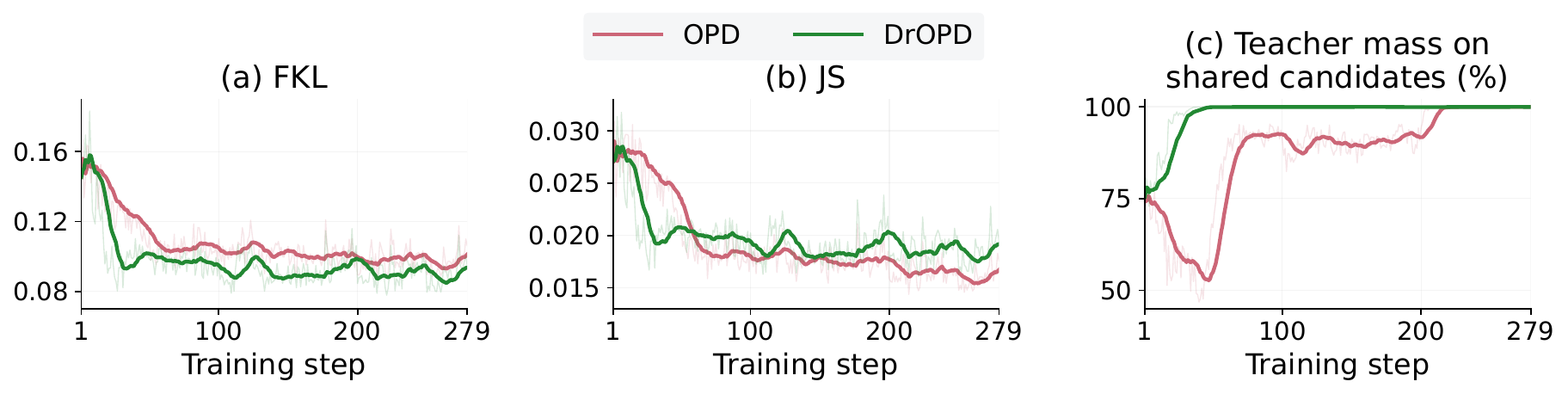}
\caption{\textbf{Additional forward-KL accelerates early alignment.} Changes in (a) FKL and (b) JS divergence during training. (c) Full-softmax teacher mass on the top-16 intersection.}
\label{fig:training_alignment}
\end{figure}

We examine how DrOPD changes alignment during training through FKL, JS divergence, and teacher probability mass on the intersection of the two top-16 candidate sets (Figure~\ref{fig:training_alignment}). With additional FKL supervision, DrOPD maintains lower FKL than OPD over most of training (Figure~\ref{fig:training_alignment}a). DrOPD also reduces JS divergence more rapidly during the early updates (Figure~\ref{fig:training_alignment}b), consistent with directing additional supervision to high-disagreement positions. This faster initial alignment is consistent with the earlier gains and higher peak in reasoning accuracy shown in Figure~\ref{fig:training_curve}c, although its JS divergence becomes slightly higher than OPD's later in training. Teacher probability mass on the top-16 intersection provides a complementary view of candidate coverage (Figure~\ref{fig:training_alignment}c). This metric sums the teacher's probabilities over the top-k shared candidates retained by both models. Under DrOPD, the mass quickly approaches 100\% and remains high, indicating that the student's top-16 candidates capture nearly all teacher probability mass. In contrast, OPD shows larger fluctuations and reaches a comparable level substantially later. These dynamics suggest that routed FKL helps earlier coverage of teacher-supported candidates while reducing the distributional discrepancy.

\subsection{Under-Coverage and Learning Progress}
\label{sec:gap_analysis}

\begin{figure}[t]
\centering
\includegraphics[width=\linewidth]{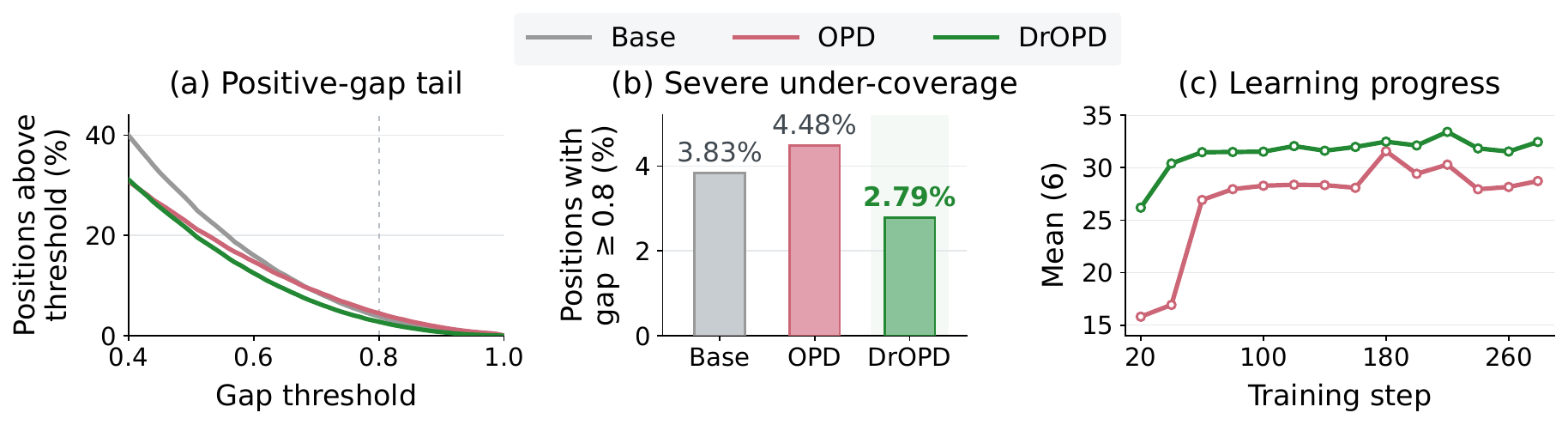}
\caption{\textbf{Under-coverage recovery and learning progress.}
(a) Fraction of fixed high-JS positions with probability gaps at or above each threshold. (b) Severe under-coverage ($\delta_t\geq0.8$) on the same positions. (c) Mean Avg@8 across six benchmarks at successive training checkpoints.}
\label{fig:undercoverage_recovery}
\label{fig:training_curve}
\end{figure}

\paragraph{Recovering under-covered teacher modes.}
We evaluate the same recorded tokens on the fixed high-JS contexts from Section~\ref{sec:undercoverage}, comparing Base, OPD, and DrOPD on the same prefixes and recorded tokens. Figure~\ref{fig:undercoverage_recovery}a reports the fraction of positions with $\delta_t\geq\tau$ as the gap threshold $\tau$ varies. DrOPD consistently reduces this proportion relative to OPD for thresholds from 0.5 to 0.9, indicating improved recovery across a range of under-coverage severities. At the severe-gap threshold of 0.8 (Figure~\ref{fig:undercoverage_recovery}b), the proportion decreases from 4.48\% under OPD to 2.79\% under DrOPD, a relative reduction of 37.9\%, and falls below the base's 3.83\%. These results support the role of routed FKL in correcting teacher-supported tokens that remain severely underweighted after RKL training.

\paragraph{Faster gains in reasoning accuracy.}
The mean accuracy in Figure~\ref{fig:undercoverage_recovery}c shows that DrOPD also learns useful task behavior earlier. Its mean accuracy of six benchmarks reaches 30.40\% at step 40, whereas OPD first exceeds 30\% at step 180. DrOPD also achieves a higher peak accuracy, showing that its gains extend beyond the early training stage. Together with the training dynamics in Section~\ref{sec:training_dynamics}, these results show that improved distributional alignment and reduced severe under-coverage accompany earlier and stronger gains in reasoning performance.

\subsection{Parameter Sensitivity}
\label{sec:ablations}

\begin{figure}[t]
\centering
\includegraphics[width=\linewidth]{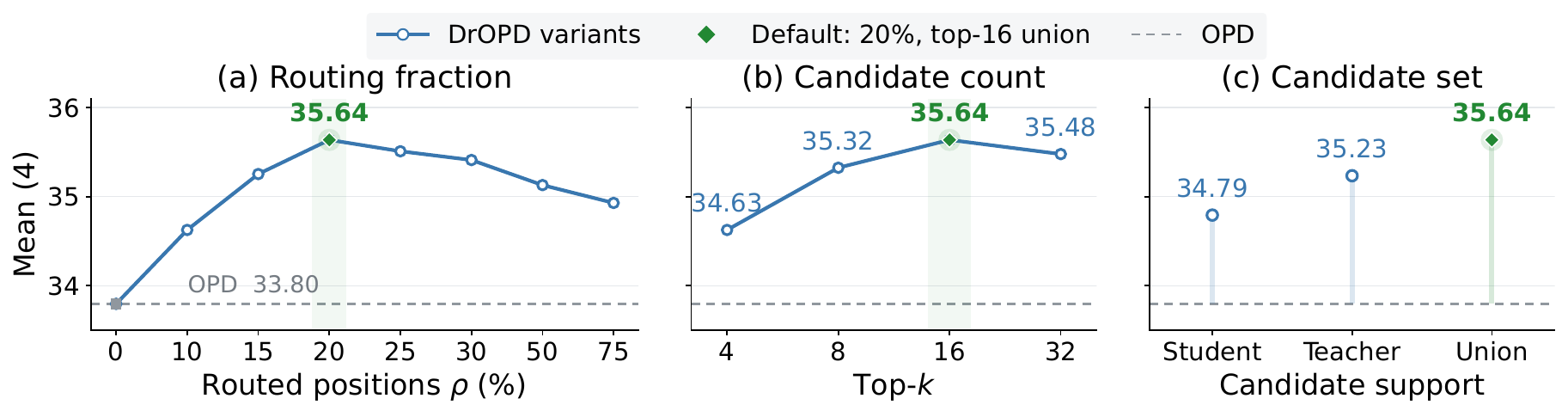}
\caption{\textbf{Sensitivity to routing and candidate support.} Avg@8 averaged over four benchmarks for (a) the routed fraction, (b) top-$k$ per model, and (c) candidate-set construction. The default uses top-20\% JS routing with a top-16 union. The dashed line denotes OPD, equivalent to $\rho=0$.}
\label{fig:ablation}
\end{figure}

We analysis the effects of the routing fraction $\rho$, the candidate count $k$, and candidate-set construction in Figure~\ref{fig:ablation}. Due to computational constraints, these experiments use four benchmarks: AIME24, AIME25, AMC23, and MATH500. Each problem samples eight responses and report mean Avg@8 across these benchmarks. All experiments use the 1.7B--4B model pair.

\paragraph{Routing fraction.}
Adding FKL supervision to the top 10\% of high-JS positions increases mean Avg@8 from 33.80\% to 34.63\%, and 20\% reaches the best, while larger routing fractions yield no further improvement (Figure~\ref{fig:ablation}a). These results favor concentrating additional supervision on a moderate fraction of high-disagreement positions rather than extending it broadly across the trajectory.

\paragraph{Candidate count.}
When $k$ is small, the truncated distribution can differ substantially from the full-vocabulary, limiting the accuracy of the loss approximation. Increasing $k$ improves this approximation. Figure~\ref{fig:ablation}b shows that performance gains saturate around $k=16$, which we choose as the default to balance accuracy and computational cost.

\paragraph{Candidate-set construction.}
Using only the student's candidates can exclude teacher-supported tokens that need FKL correction, while using only the teacher's candidate may omit student-preferred tokens relevant to RKL alignment. Their union retains both groups, supporting both RKL alignment and FKL correction. As shown in Figure~\ref{fig:ablation}c, the union achieves the highest mean Avg@8 of 35.64\%, compared with 34.79\% for student-only and 35.23\% for teacher-only selection.

\subsection{Loss Direction and Supervision Allocation}
\label{sec:loss_allocation}

Our cross-sampling results suggest that a small subset of high-disagreement positions accounts for much of the benefit of teacher intervention. This raises a natural question: \emph{can distillation supervision also be restricted to these positions?} To investigate this, we rank positions by teacher--student JS divergence and divide them into the bottom 80\% and top 20\%. We then apply different combinations of RKL and FKL to these two regions, as summarized in Table~\ref{tab:loss_ablation}. All variants use the same top-16 union candidate set and are evaluated by mean Avg@8 across the four benchmarks.

Full RKL (standard OPD) and Full FKL apply the respective loss to both regions, whereas Sparse RKL and Sparse FKL supervise only the top-20\% high-JS positions. DrOPD retains RKL in both regions and adds FKL only to the top 20\%. Sparse RKL improves mean Avg@8 from 33.80\% to 34.94\%, supporting the value of concentrating supervision at high-disagreement positions. However, the same restriction reduces FKL performance from 34.16\% to 33.29\%. Thus, the benefit of sparse supervision depends on the loss direction. DrOPD achieves the highest mean Avg@8 of 35.64\%. Together, these results favor combining broad RKL alignment with selective FKL supervision over applying either objective alone in the tested configurations.

The strong Sparse RKL also raises a further question: could two directions benefit from different position or allocations that change during training? DrOPD provides a simple effective allocation, while jointly adapting the placement and relative strength of RKL and FKL remains for future work.

\begin{table}[t]
\centering\small\setlength{\tabcolsep}{6pt}
\caption{\textbf{Effect of loss direction and supervision allocation.}. Selected positions are the top 20\% ranked by JS; remaining positions are the other 80\%. ``--'' denotes no loss at those positions.}
\label{tab:loss_ablation}
\begin{tabular}{lccr}
\toprule
Method & Remaining positions & Selected positions & Mean(4)\\
\midrule
OPD (Full RKL) & RKL & RKL & 33.80 \\
Full FKL & FKL & FKL & 34.16 \\
Sparse RKL & -- & RKL & 34.94 \\
Sparse FKL & -- & FKL & 33.29 \\
\midrule
\rowcolor{mygreen!12} DrOPD & RKL & RKL + FKL & \textbf{35.64} \\
\bottomrule
\end{tabular}

\end{table}

\section{Related Work}
\label{sec:related_work}

\paragraph{On-policy distillation and its mechanisms.}
OPD trains a student on its own generated trajectories using teacher feedback, reducing the mismatch between training contexts and those encountered at inference. GKD provides a framework for learning from student-generated sequences with different divergence objectives~\citep{agarwal2023onpolicy}, while MiniLLM develops RKL-based distillation to suppress student probability in regions that the teacher considers unlikely~\citep{gu2024minillm}. ExOPD generalizes this formulation through reward extrapolation~\citep{yang2026learningbeyond}. Recent studies examine when this supervision succeeds, highlighting teacher--student compatibility, teacher reliability on student prefixes, and the effects of divergence estimation~\citep{li2026rethinking,fu2026revisiting,zhu2026manyfaces}. Corresponding improvements include more informative state exploration and calibrated teacher guidance in Uni-OPD~\citep{hou2026uni}, variance reduction through control variates in vOPD~\citep{oh2026klbaseline}, and explicit treatment of omitted probability mass in Tail-Aware Top-$k$ OPD~\citep{huang2026tailaware}. Self-distillation constructs a teacher from the model itself: SDPO conditions the model on textual feedback~\citep{hubotter2026reinforcement}, while Self-Distilled Reasoner uses verified reasoning traces as privileged context~\citep{zhao2026selfdistilled}. Our work focuses on a fixed external teacher and studies how to supplement RKL at positions where teacher-supported tokens remain severely under-covered.

\paragraph{Token importance and selective supervision.}
Selective distillation allocates supervision according to the usefulness or reliability of individual tokens. REOPOLD combines entropy-based token sampling with reward clipping~\citep{ko2026scaling}, and SelecTKD uses teacher verification to retain or down-weight token-level distillation losses~\citep{selectkd2025}. FiRe-OPD filters trajectories using teacher likelihood and then reweights tokens using teacher confidence and student uncertainty~\citep{li2026fireopd}. TIP jointly analyzes student entropy and teacher--student divergence, identifying low-entropy, high-disagreement positions that entropy-only selection can miss~\citep{xu2026tip}. In self-distillation, PW-OPSD uses position-dependent weighting to account for teacher reliability along reasoning trajectories~\citep{liu2026teacher}, while DASH uses disagreement to adapt the temporal extent of supervision~\citep{hou2026dash}. \citet{meng2026sparse} use cross-sampling between base and RL-trained models to show that sparse high-divergence positions account for much of the performance change after RLVR. Building on this diagnostic, we compare routing signals through their downstream effects at comparable teacher-intervention ratios. We further examine the probability deficits at high-JS positions on fixed contexts.

\paragraph{Divergence objectives and teacher-mode recovery.}
Several methods adapt the form of supervision to address limitations of a uniform distillation objective. EOPD retains RKL and adds FKL where teacher entropy exceeds a threshold~\citep{jin2026entropy}. AOPD retains reinforcement updates for positive-advantage tokens and replaces non-positive-advantage updates with imitation over teacher top-$k$ candidates~\citep{jia2026asymmetric}. TIDE independently selects student-excess and student-deficit positions, applying bounded Hellinger corrections to the former and teacher top-$k$ supervision to recover missing probability mass at the latter~\citep{yu2026mismatch}. TRACE applies FKL to annotated key reasoning spans, optionally applies RKL to error spans, and combines distillation with GRPO; it also supports online self-annotation~\citep{trace2026}. In privileged-context self-distillation, DASD uses entropy to choose between attraction toward and repulsion from the self-teacher~\citep{zhang2026tailoring}. DrOPD shares the motivation of recovering under-covered teacher modes, but uses a different supervision allocation: all positions retain RKL, and the highest-JS positions receive additional FKL. Our contribution connects this allocation to cross-sampling evidence of behavioral importance and fixed-context evidence of severe under-coverage, without requiring reasoning-span annotations.

% \begin{figure}[t]
%     \centering
%     \includegraphics[width=1.0\linewidth]{fig/passk.pdf}
%     \caption{
%     Pass@$k$ performance across different methods
%     }
% \label{fig:passk}
% \end{figure}

\section{Conclusion}

We study where standard on-policy distillation fails to effectively transfer teacher knowledge. Controlled cross-sampling shows that teacher--student distributional disagreement identifies behaviorally critical positions more effectively than single-model uncertainty, while further analysis reveals severe teacher-mode under-coverage at many of these positions. Motivated by this diagnosis, we propose Divergence-Routed On-Policy Distillation (DrOPD), which retains RKL supervision throughout student-generated trajectories and selectively adds FKL supervision to the highest-disagreement positions. Experiments across multiple mathematical reasoning benchmarks and student--teacher configurations show that DrOPD consistently improves over standard OPD and entropy-based routing strategies. Our analyses further support targeted mode-covering supervision as an effective way to address teacher knowledge that is difficult to recover with RKL alone.

\bibliography{iclr2027_conference}
\bibliographystyle{iclr2027_conference}

\clearpage
\appendix
\section{Experimental Details}
\label{app:exp_details}

\subsection{Models, Data, and Optimization}
We use verl~\citep{sheng2024hybridflow} to train the two student--teacher pairs described in Section 4, keeping each teacher frozen. Students train on processed DAPO-Math-17K prompts for one epoch with four responses per prompt. Table~\ref{tab:app_settings} summarizes the settings. Student advantages use the distillation signal without adding answer-correctness rewards.

\begin{table}[htbp]
\centering\small
\caption{Training and evaluation settings. Candidate top-$k$ specifies the support used for distillation, rather than a restriction on rollout sampling.}
\label{tab:app_settings}
\begin{tabular}{ll}
\toprule
Setting & Value \\
\midrule
Training framework / rollout engine & verl / vLLM \\
Training data & DAPO-Math-17K \\
Training epochs & 1 \\
Prompt batch / PPO minibatch size & 64 / 64 \\
Responses per training prompt & 4 \\
Learning rate / weight decay & $10^{-6}$ / 0.01 \\
Learning-rate warmup fraction & 0 \\
Maximum prompt / response length & 1,024 / 7,168 tokens \\
Rollout / teacher temperature & 1.0 / 1.0 \\
Candidate support & Top-16 union \\
Routed fraction / added FKL coefficient & 20\% / 1 \\
PPO epochs / ratio clipping range & 1 / $[0.8,1.2]$ \\
Loss aggregation & Candidate sum, valid-response-token mean \\
Number of GPUs & 8*A800 \\
\midrule
Benchmark responses per problem & 8 \\
Benchmark temperature / top-$p$ & 0.7 / 0.95 \\
Benchmark response limit & 7,168 tokens \\
\bottomrule
\end{tabular}
\end{table}

\subsection{Prompting and Evaluation}
Evaluation and cross-sampling append ``Please reason step by step, and put your final answer within \texttt{\textbackslash boxed\{\}}.'' to each problem. We apply the tokenizer's chat template with a generation prompt and \texttt{enable\_thinking=False}. The evaluator checks the extracted boxed answer by normalized matching or symbolic equivalence; responses without an extractable boxed answer are incorrect.

For a benchmark $\mathcal D$ with $n$ problems, let $b_{ij}\in\{0,1\}$ indicate whether response $j$ to problem $i$ is correct. We report
\begin{equation}
\operatorname{Avg@8}(\mathcal D)=\frac{1}{8n}\sum_{i=1}^{n}\sum_{j=1}^{8}b_{ij},\qquad
\operatorname{Mean}(m)=\frac{1}{m}\sum_{d=1}^{m}\operatorname{Avg@8}(\mathcal D_d).
\label{eq:app_avg8}
\end{equation}
Benchmarks have equal weight in each mean. Mean (6) includes AIME24, AIME25, AMC23, Minerva, Olympiad, and MATH500. Mean (4), used for sensitivity and loss allocation, includes AIME24, AIME25, AMC23, and MATH500.

\subsection{Training Cost}
Table~\ref{tab:app_training_cost} reports the logged training cost for the 1.7B--4B student--teacher pair over 279 steps. Mean step time and cumulative step time are computed from the per-step training timers; cumulative time excludes initialization, checkpoint saving, and offline evaluation. Mean response length is averaged over the logged training steps.

\begin{table}[htbp]
\centering\small
\caption{Training cost for the 1.7B--4B pair over 279 steps.}
\label{tab:app_training_cost}
\begin{tabular}{llrrr}
\toprule
Model pair & Method & \shortstack{Mean step time\\(s)} & \shortstack{Cumulative step time\\(h)} & \shortstack{Mean response length\\(tokens)} \\
\midrule
1.7B--4B & OPD & 141.43 & 10.96 & 3117 \\
1.7B--4B & DrOPD & 152.62 & 11.83 & 3733 \\
\bottomrule
\end{tabular}
\end{table}

\subsection{Baseline Configurations}
OPD uses all-position RKL on the top-16 union. The entropy baselines retain RKL and add FKL at the top-20\% positions ranked by student or teacher entropy on the normalized union. EOPD adds FKL where teacher entropy exceeds 0.8. Each added FKL term has coefficient one. Full FKL supervises all positions; Sparse RKL and Sparse FKL supervise only the top-20\% JS positions. Candidate-set ablations retain the DrOPD allocation and change only the candidate support.

\clearpage
\section{Algorithm and Gradient Analysis}
\label{app:implementation}

\subsection{Training Procedure}
Algorithm~\ref{alg:app_dropd} summarizes DrOPD. Each position receives a distributional learning signal over its candidate set, even though the rollout records only one sampled token. Consequently, the added FKL can assign supervision to teacher-supported candidates that were not sampled at that position.

\begin{algorithm}[htbp]
\caption{Divergence-Routed On-Policy Distillation}
\label{alg:app_dropd}
\begin{algorithmic}[1]
\Require Student $\pi_\theta$, frozen teacher $\pi_{\mathrm{tch}}$, prompts $\mathcal D$, $k=16$, $\rho=0.2$
\For{each prompt batch from $\mathcal D$}
    \State Set $\pi_{\mathrm{old}}\gets\pi_\theta$ and generate four responses per prompt
    \State Form contexts $c_t$ and the valid response-position set $\mathcal B$
    \For{each $t\in\mathcal B$}
        \State Obtain student and teacher top-$k$ candidates on $c_t$
        \State Form their union $\mathcal C_t$ and mask duplicate candidate IDs
        \State Normalize both distributions on $\mathcal C_t$ to obtain $\widetilde p_t,\widetilde q_t$
        \State Compute $d_t=D_{\mathrm{JS}}(\widetilde p_t,\widetilde q_t)$
    \EndFor
    \State Determine the top-$\rho$ cutoff $\epsilon_\rho$ over valid positions
    \State Set $g_t=\mathds{1}[d_t\geq\epsilon_\rho]$ for each valid position
    \State Construct all-position RKL rewards and add the FKL term where $g_t=1$
    \State Detach the candidate advantages as in Equation 9
    \State Update $\theta$ with the clipped surrogate in Equation 10
\EndFor
\end{algorithmic}
\end{algorithm}

The union contains at most $2k$ distinct candidates. Candidate IDs, routing decisions, and advantage coefficients remain fixed during each policy update; gradients flow through the student policy ratios. Candidate contributions are summed at each position and the resulting objective is averaged over valid response positions. Prompt tokens and padding do not contribute to this average. We use one PPO epoch and one minibatch per on-policy update.

The percentile router operates over the valid positions available to an actor reward-computation call. It sets $K_\rho=\operatorname{round}(\rho|\mathcal B|)$ and uses the $K_\rho$-th largest score as the cutoff. Ties at the cutoff are included, so the selected fraction can slightly exceed $\rho$. An empty population or $K_\rho=0$ yields no selected positions. Selection is performed over the batch population rather than independently within each response. Routing controls the additional FKL signal; RKL remains active at every valid position.

\subsection{Derivation of the KL Logit Gradients}
Let $p(v)=\operatorname{softmax}(z)(v)$ on a fixed support $\mathcal V$, and let $q$ be a fixed positive teacher distribution on that support. The softmax Jacobian is $\partial p(u)/\partial z(v)=p(u)(\mathds{1}[u=v]-p(v))$. Applying the chain rule gives
\begin{align}
\frac{\partial D_{\mathrm{KL}}(p\|q)}{\partial z(v)}
&=\sum_{u\in\mathcal V}p(u)(\mathds{1}[u=v]-p(v))\left(\log\frac{p(u)}{q(u)}+1\right)\nonumber\\
&=p(v)\left(\log\frac{p(v)}{q(v)}-D_{\mathrm{KL}}(p\|q)\right),
\label{eq:app_rkl_gradient}\\
\frac{\partial D_{\mathrm{KL}}(q\|p)}{\partial z(v)}
&=-\sum_{u\in\mathcal V}q(u)(\mathds{1}[u=v]-p(v))=p(v)-q(v).
\label{eq:app_fkl_gradient}
\end{align}
For a finite support and fixed positive $q$, the RKL is bounded above by $\log(1/\min_u q(u))$. Since $p(v)\log p(v)\to0$ as $p(v)\to0$, the RKL gradient in Equation~\ref{eq:app_rkl_gradient} vanishes in this limit. The FKL gradient instead converges to $-q(v)$, retaining a corrective signal that increases the student's logit under gradient descent. These identities describe the local gradient behavior at a fixed context and motivate the complementary supervision; the empirical analyses evaluate its effect during training. They also apply to normalized candidate distributions when the support is held fixed.

\clearpage
\subsection{From Distillation Loss to Candidate Rewards}
\label{app:reward_derivation}
We derive the candidate rewards by matching the negative gradient of the distillation loss to a weighted sum of student log-probability gradients. Fix the context $c_t$, the candidate set $\mathcal C_t$, the teacher distribution $\widetilde q_t$, and the routing gate $g_t$. The normalized student distribution $\widetilde p_t$ depends on $\theta$. The per-position loss in Equation~(8) of the main paper is
\begin{equation}
\ell_t(\theta)=\ell_{\mathrm{RKL},t}+g_t\ell_{\mathrm{FKL},t}
=\sum_{v\in\mathcal C_t}\widetilde p_t(v)\log\frac{\widetilde p_t(v)}{\widetilde q_t(v)}
+g_t\sum_{v\in\mathcal C_t}\widetilde q_t(v)\log\frac{\widetilde q_t(v)}{\widetilde p_t(v)}.
\label{eq:app_reward_loss}
\end{equation}
Our goal is to identify coefficients $R_t(v)$ such that $-\nabla_\theta\ell_t=\sum_{v\in\mathcal C_t}R_t(v)\nabla_\theta\log\widetilde p_t(v)$. The coefficients are treated as detached weights when applying this gradient; they need not equal the negative loss value.

\paragraph{RKL contribution.}
Both the weighting distribution and the logarithm in RKL depend on the student parameters. Differentiating and using $\nabla_\theta\widetilde p_t(v)=\widetilde p_t(v)\nabla_\theta\log\widetilde p_t(v)$ gives
\begin{align}
\nabla_\theta\ell_{\mathrm{RKL},t}
&=\sum_{v\in\mathcal C_t}\left(\log\frac{\widetilde p_t(v)}{\widetilde q_t(v)}+1\right)\nabla_\theta\widetilde p_t(v)\nonumber\\
&=\sum_{v\in\mathcal C_t}\widetilde p_t(v)\left(\log\frac{\widetilde p_t(v)}{\widetilde q_t(v)}+1\right)\nabla_\theta\log\widetilde p_t(v).
\label{eq:app_reward_rkl_chain}
\end{align}
The constant contribution cancels because the student probabilities sum to one on the fixed candidate set:
\begin{equation}
\sum_{v\in\mathcal C_t}\widetilde p_t(v)\nabla_\theta\log\widetilde p_t(v)
=\nabla_\theta\sum_{v\in\mathcal C_t}\widetilde p_t(v)=0.
\label{eq:app_reward_normalization}
\end{equation}
Therefore,
\begin{equation}
-\nabla_\theta\ell_{\mathrm{RKL},t}
=\sum_{v\in\mathcal C_t}\underbrace{\left[-\widetilde p_t(v)\log\frac{\widetilde p_t(v)}{\widetilde q_t(v)}\right]}_{R_{\mathrm{RKL},t}(v)}\nabla_\theta\log\widetilde p_t(v).
\label{eq:app_reward_rkl}
\end{equation}
Thus, the RKL reward coefficient is the negative per-candidate contribution to RKL, including its student-probability weight.

\paragraph{FKL contribution.}
The teacher distribution is fixed, so FKL decomposes into a constant teacher term and a student cross-entropy term:
\begin{equation}
\ell_{\mathrm{FKL},t}
=\underbrace{\sum_{v\in\mathcal C_t}\widetilde q_t(v)\log\widetilde q_t(v)}_{\text{independent of }\theta}
-\sum_{v\in\mathcal C_t}\widetilde q_t(v)\log\widetilde p_t(v).
\end{equation}
Only the second term contributes to the gradient, yielding
\begin{equation}
-\nabla_\theta\ell_{\mathrm{FKL},t}
=\sum_{v\in\mathcal C_t}\underbrace{\widetilde q_t(v)}_{R_{\mathrm{FKL},t}(v)}\nabla_\theta\log\widetilde p_t(v).
\label{eq:app_reward_fkl}
\end{equation}
The corresponding coefficient is therefore simply the teacher probability, rather than the per-candidate FKL value. Combining the two contributions gives
\begin{equation}
\boxed{R_t(v)=-\widetilde p_t(v)\log\frac{\widetilde p_t(v)}{\widetilde q_t(v)}+g_t\widetilde q_t(v),\qquad v\in\mathcal C_t.}
\label{eq:app_reward_combined}
\end{equation}
This recovers Equation~(9) of the main paper through the negative-gradient identity above. The different forms of the RKL and FKL rewards follow from differentiating a student-weighted divergence versus a teacher-weighted divergence, rather than from using different definitions of reward.

\clearpage
\section{Cross-Sampling Protocol and Additional Results}
\label{app:cross_sampling}

\subsection{Generation and Routing}
We evaluate 40 AMC23 problems with eight responses each, giving 320 responses per routing configuration. Both models remain frozen. The default sampler is the base student, and the teacher supplies the next token only when the routing condition is satisfied. Both models subsequently condition on the same updated prefix, including any teacher-sampled token. Routing divergences use temperature-one distributions normalized on the top-16 union. Token generation uses temperature 0.7, a 7,168-token response limit, and full-vocabulary sampling without top-$p$ or top-$k$ truncation. The top-16 approximation therefore controls the routing score, not the set of tokens that can be generated.

For divergence and entropy criteria, teacher intervention occurs when the score exceeds the corresponding threshold. For random routing, each position is assigned to the teacher with probability $r$, independently of the distributions. Tables~\ref{tab:app_cross_divergence} and~\ref{tab:app_cross_entropy} give the numerical threshold sweeps underlying Figure 1a, including points outside its displayed range. Entropy-only cross-sampling uses the indicated model's full next-token distribution. All entropies and divergences use natural logarithms.

The intervention rate $\eta$ is the mean of the per-response intervention fractions, as defined in Equation 2. We use the observed rate rather than the nominal threshold or random-routing probability when comparing strategies. Different routing policies can produce different prefixes and response lengths; they are evaluated on the same problems, but do not share a fixed set of decoding contexts. Standalone Base and Teacher achieve 14.38\% and 60.63\% Avg@8, respectively.

\begin{table}[!htbp]
\centering\setlength{\tabcolsep}{5pt}
\caption{Divergence-based cross-sampling on AMC23. $\tau$ is the routing threshold; $\eta$ is the measured intervention rate. Avg@8 is reported in percent.}
\label{tab:app_cross_divergence}
\begin{tabular}{rrr@{\hspace{14pt}}rrr@{\hspace{14pt}}rrr}
\toprule
\multicolumn{3}{c}{JS} & \multicolumn{3}{c}{RKL} & \multicolumn{3}{c}{FKL} \\
\midrule
$\tau$ & $\eta$ (\%) & Avg@8 & $\tau$ & $\eta$ (\%) & Avg@8 & $\tau$ & $\eta$ (\%) & Avg@8 \\
\midrule
0.5 & 0.21 & 37.81 & 3 & 0.70 & 40.31 & 2.2 & 0.54 & 36.56 \\
0.4 & 0.59 & 40.62 & 2.6 & 0.97 & 43.12 & 1.9 & 0.79 & 40.62 \\
0.3 & 1.48 & 45.94 & 2.2 & 1.33 & 46.88 & 1.6 & 1.22 & 44.06 \\
0.2 & 3.24 & 49.06 & 1.8 & 1.97 & 45.62 & 1 & 2.80 & 46.25 \\
0.18 & 3.93 & 51.88 & 1.4 & 3.11 & 50.62 & 0.7 & 4.41 & 52.50 \\
0.16 & 4.46 & 54.06 & 1 & 4.89 & 56.88 & 0.6 & 5.16 & 54.06 \\
0.12 & 6.08 & 56.88 & 0.8 & 6.34 & 58.13 & 0.5 & 6.10 & 54.37 \\
0.1 & 7.24 & 55.94 & 0.7 & 6.99 & 58.44 & 0.4 & 7.13 & 57.19 \\
0.08 & 8.69 & 57.81 & 0.6 & 8.11 & 60.62 & 0.3 & 8.77 & 63.12 \\
0.07 & 9.23 & 60.31 & 0.5 & 9.30 & 64.06 & 0.2 & 11.08 & 63.44 \\
0.06 & 10.13 & 65.00 & 0.4 & 10.83 & 62.50 & 0.1 & 14.29 & 58.75 \\
0.05 & 11.22 & 61.88 & 0.3 & 12.95 & 61.25 & 0.08 & 15.69 & 61.56 \\
0.04 & 12.87 & 61.88 & 0.2 & 15.49 & 65.00 & 0.06 & 17.23 & 61.56 \\
0.02 & 16.19 & 65.31 & 0.1 & 19.32 & 62.81 & 0.05 & 18.00 & 62.81 \\
0.01 & 19.65 & 63.44 & -- & -- & -- & -- & -- & -- \\
\bottomrule
\end{tabular}

\end{table}

\clearpage
\begin{table}[!htbp]
\centering\setlength{\tabcolsep}{5pt}
\caption{Entropy-based and random cross-sampling on AMC23. $r$ is the random teacher-intervention probability; the remaining columns follow Table~\ref{tab:app_cross_divergence}.}
\label{tab:app_cross_entropy}
\begin{tabular}{rrr@{\hspace{14pt}}rrr@{\hspace{14pt}}rrr}
\toprule
\multicolumn{3}{c}{Student entropy} & \multicolumn{3}{c}{Teacher entropy} & \multicolumn{3}{c}{Random} \\
\midrule
$\tau$ & $\eta$ (\%) & Avg@8 & $\tau$ & $\eta$ (\%) & Avg@8 & $r$ & $\eta$ (\%) & Avg@8 \\
\midrule
1.8 & 0.82 & 34.69 & 1.6 & 3.83 & 25.00 & 0.04 & 3.67 & 16.56 \\
1.6 & 1.30 & 37.19 & 1.4 & 4.51 & 27.81 & 0.06 & 5.36 & 16.88 \\
1.4 & 2.03 & 40.00 & 1.2 & 5.42 & 29.69 & 0.08 & 7.30 & 17.50 \\
1.2 & 3.12 & 42.50 & 1 & 6.28 & 37.50 & 0.1 & 9.03 & 20.00 \\
1 & 4.79 & 42.19 & 0.9 & 6.78 & 31.25 & 0.12 & 10.95 & 20.31 \\
0.9 & 5.84 & 45.94 & 0.8 & 7.90 & 34.06 & 0.14 & 12.88 & 20.94 \\
0.8 & 6.80 & 48.44 & 0.7 & 9.01 & 37.19 & 0.18 & 16.50 & 24.38 \\
0.7 & 9.00 & 50.00 & 0.6 & 11.04 & 40.94 & 0.2 & 18.39 & 26.56 \\
0.6 & 11.03 & 50.62 & 0.5 & 12.31 & 35.31 & 0.22 & 20.19 & 27.81 \\
0.5 & 12.51 & 57.19 & 0.4 & 13.94 & 40.94 & 0.24 & 22.18 & 26.56 \\
0.4 & 14.53 & 55.62 & 0.3 & 15.20 & 41.88 & -- & -- & -- \\
0.3 & 16.56 & 56.88 & 0.2 & 15.44 & 45.00 & -- & -- & -- \\
0.2 & 19.62 & 60.00 & 0.1 & 17.99 & 45.62 & -- & -- & -- \\
\bottomrule
\end{tabular}

\end{table}

\subsection{Entropy Regions within High-JS Positions}
We further restrict teacher intervention to one of four entropy regions within positions with JS $>0.06$. Student and teacher entropy are each computed on the corresponding model's normalized top-16 distribution. High and low entropy denote values above and below 0.5, respectively. The region condition is evaluated on the current prefix at every decoding step, together with the JS condition.

\begin{table}[htbp]
\centering\small
\caption{Cross-sampling restricted to individual entropy regions. Each row is a separate policy on the same 40 AMC23 problems. The column order explicitly distinguishes student and teacher entropy.}
\label{tab:app_entropy_regions}
\begin{tabular}{llrr}
\toprule
Student entropy & Teacher entropy & Intervention (\%) & Avg@8 (\%) \\
\midrule
High & High & 7.19 & 27.81 \\
High & Low & 3.82 & 47.50 \\
Low & High & 1.20 & 18.75 \\
Low & Low & 0.16 & 16.25 \\
\bottomrule
\end{tabular}
\end{table}

Table~\ref{tab:app_entropy_regions} shows that intervention is particularly useful when the student is uncertain and the teacher has a concentrated preference: the high-student/low-teacher region reaches 47.50\% Avg@8 at 3.82\% intervention. This provides a complementary description of high-disagreement positions. Because the four policies have different realized intervention rates and induce different trajectories, the table does not isolate a causal effect of entropy at a matched budget. DrOPD uses JS directly and does not require an entropy-region classifier.

\clearpage
\section{Fixed-Context Under-Coverage Analysis}
\label{app:distribution}

\subsection{Shared Trajectories and Recorded Tokens}
The fixed-context analysis uses 320 Base--Teacher cross-sampled responses generated with the JS threshold 0.06. The shared archive contains 996,919 response positions, of which 97,122 satisfy the original JS $>0.06$ condition. We hold the prefixes, recorded token IDs, and this original high-JS mask fixed when comparing Base, OPD, and DrOPD. At the selected positions, the recorded token was sampled from the teacher. We check alignment of example IDs, response indices, token IDs, routing decisions, and stored teacher probabilities across the model-specific records.

Each position contributes the probability of its actually recorded token, rather than a maximum gap over candidate tokens. The replay cache stores probabilities normalized over each model's retained top-16 candidates, assigning zero when the recorded token is outside that set. We compute $\delta_t=[q_t(v_t)-p_t(v_t)]_+$ from these stored probabilities, following the positive-gap definition in Equation 3. The analysis compares how different students score the same teacher-generated tokens on the same contexts.

\subsection{Conditional Means and Tail Fractions}
Let $\mathcal D_j$ be the $j$-th decile of positions ranked by their original Base--Teacher JS scores. Figure 1b reports the conditional positive-gap mean
\begin{equation}
\mu_j^+=\frac{\sum_{t\in\mathcal D_j}\delta_t\mathds{1}[\delta_t>0]}{\sum_{t\in\mathcal D_j}\mathds{1}[\delta_t>0]}.
\label{eq:app_positive_mean}
\end{equation}
Positions with zero positive gap are excluded from this mean. Deciles are determined once using the original JS scores; the highest decile begins at approximately 0.05746. It therefore closely overlaps, but is not identical to, the fixed JS $>0.06$ subset used for the tail comparisons.

Let $\mathcal H=\{t:d_t^{\mathrm{Base}}>0.06\}$ denote this fixed high-JS subset. For model $M$ and a positive threshold $\tau$, define
\begin{equation}
F_M(\tau)=\frac{1}{|\mathcal H|}\sum_{t\in\mathcal H}\mathds{1}[\delta_t^M\geq\tau],\qquad
C_{\mathrm{OPD}}(\tau)=100\left(\frac{F_{\mathrm{OPD}}(\tau)}{F_{\mathrm{Base}}(\tau)}-1\right).
\label{eq:app_gap_tail}
\end{equation}
Figure 4a plots $100F_M(\tau)$, while Figure 1c plots $C_{\mathrm{OPD}}(\tau)$ wherever the Base tail is nonzero. Each tail fraction uses all positions in $\mathcal H$ as its denominator, including positions without a positive gap. The relative-change curve measures percentage change, rather than a difference in percentage points. The severe-gap bars in Figure 4b are the values at $\tau=0.8$.

\begin{table}[htbp]
\centering\small
\caption{Fraction (\%) of fixed original high-JS positions above each recorded-token gap threshold. Every row uses the same 97,122 positions.}
\label{tab:app_gap_tails}
\begin{tabular}{rrrr}
\toprule
Gap threshold $\tau$ & Base & OPD & DrOPD \\
\midrule
0.2 & 73.01 & 51.40 & 56.86 \\
0.4 & 39.97 & 30.82 & 31.16 \\
0.5 & 26.35 & 22.05 & 20.68 \\
0.6 & 16.03 & 14.78 & 12.48 \\
0.7 & 8.72 & 8.88 & 6.57 \\
\rowcolor{mygreen!12} 0.8 & 3.83 & 4.48 & 2.79 \\
0.9 & 1.12 & 1.65 & 0.71 \\
\bottomrule
\end{tabular}

\end{table}

Table~\ref{tab:app_gap_tails} shows how the comparison changes with deficit severity. Both trained students reduce moderate gaps, while DrOPD also reduces the extreme tail. At threshold 0.8, the fractions are 3.83\% for Base, 4.48\% for OPD, and 2.79\% for DrOPD. The relative reduction from OPD is computed from unrounded fractions. The selected positions comprise 9.74\% of all archived tokens; this token-pooled fraction differs from the mean per-response intervention rate used in the cross-sampling curves.

\clearpage
\section{Training Diagnostic Definitions}
\label{app:training_logs}

\subsection{Distributional Alignment}
The FKL and JS curves in Figure 3 summarize each method's own student-generated training rollouts. At every valid response position, we form the deduplicated top-16 union and normalize the student and teacher distributions on that support. The logged values average the corresponding divergence over all valid positions, including positions not selected for additional FKL. This differs from the fixed-context analysis in Appendix~\ref{app:distribution}, where the prefixes and recorded tokens are shared across models.

The distributions are evaluated using natural logarithms. The displayed training curves show raw step values together with trailing means over at most 15 steps. For a metric $a_s$ at training step $s$, the smoothed value is $\overline a_s=L_s^{-1}\sum_{j=s-L_s+1}^{s}a_j$, where $L_s=\min(15,s)$. Benchmark Mean (6) in Figure 4c is computed from the eight-response evaluation results at each available evaluation step and is plotted without smoothing. It is an offline reasoning-accuracy measure, rather than the logged correctness of a training batch.

\subsection{Probability Mass on the Top-16 Intersection}
\begin{figure}[htbp]
\centering
\includegraphics[width=.82\linewidth]{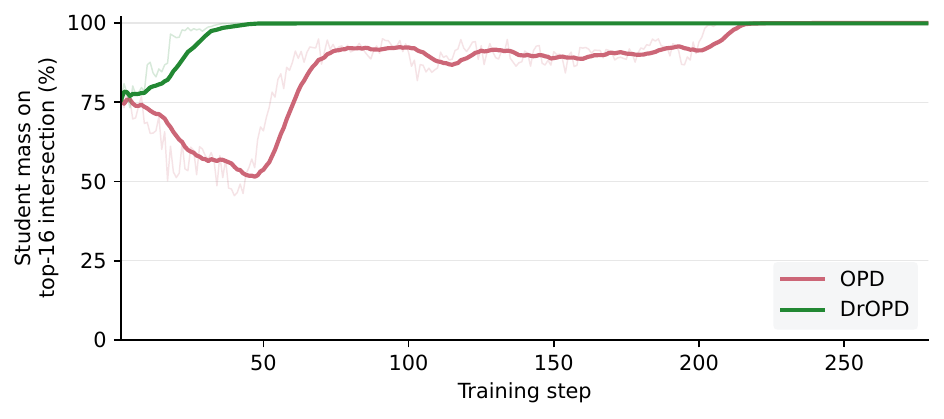}
\caption{\textbf{Student probability mass on the top-16 intersection.} Faint lines show raw step values and solid lines show trailing 15-step means. This complements the teacher-mass statistic in Figure 3c using the corresponding student probabilities.}
\label{fig:app_student_mass}
\end{figure}

Let $\mathcal I_t=\operatorname{Top16}(p_t)\cap\operatorname{Top16}(q_t)$ and let $\mathcal B_+=\{t\in\mathcal B:|\mathcal I_t|>0\}$. Using full-softmax model probabilities, the teacher and student intersection masses are
\begin{equation}
M_q=\frac{100}{|\mathcal B_+|}\sum_{t\in\mathcal B_+}\sum_{v\in\mathcal I_t}\pi_{\mathrm{tch}}(v\mid c_t),\qquad
M_p=\frac{100}{|\mathcal B_+|}\sum_{t\in\mathcal B_+}\sum_{v\in\mathcal I_t}\pi_\theta(v\mid c_t).
\label{eq:app_intersection_mass}
\end{equation}
These probabilities are not renormalized on the intersection. Positions with an empty intersection are excluded from the logged average. Figure~\ref{fig:training_alignment}c reports $M_q$, and Figure~\ref{fig:app_student_mass} provides the complementary student quantity $M_p$. The fields are collected on training rollouts even though their logging names begin with \texttt{val-topk}.

High teacher intersection mass means that most teacher probability lies on candidates also included in the student's top-16 set. High student intersection mass indicates the analogous property for student probability. These statistics measure shared candidate support; FKL and JS additionally assess how probability is distributed within the union. In particular, candidate membership alone does not establish that the two models assign equal probability to those candidates.

DrOPD establishes high student intersection mass earlier and maintains it through the middle of training, while OPD reaches a similar level later. Together with the teacher-mass curve, this pattern supports the interpretation that additional FKL accelerates the recovery of shared candidate support.

\clearpage
\section{Disclosure of AI Use}
\label{app:ai_disclosure}

AI tools were used to assist with literature searches and the exploration of potential research directions. After the human researchers designed the method, conducted the experiments, and obtained the results, AI tools assisted with drafting manuscript sections, analyzing and visualizing existing experimental results, and preparing mathematical derivations. The manuscript was refined through iterative human review and revision, with AI assistance for language polishing. The authors take responsibility for the final content of the paper, including its claims, analyses, and conclusions.

% \appendix
% \section{Appendix}

\end{document}